\documentclass{article}

\usepackage[T1]{fontenc}
\usepackage{iclr2027_conference,times}
\iclrfinalcopy
\usepackage{amsmath,amssymb,mathtools}
\usepackage{graphicx}
\usepackage{microtype}
\usepackage[hidelinks]{hyperref}
\hypersetup{pdftitle={Predictive Closure: A Set-Level Characterization of JEPA's Feature Selection},pdfauthor={Mosh Levy, Aly Lidayan, Yoav Goldberg}}
\usepackage{url}
\usepackage{needspace}
\usepackage{placeins}
\usepackage{float}

\title{\raggedright Predictive Closure: A Set-Level Characterization of JEPA's Feature Selection}
\author{Mosh Levy \\
Bar-Ilan University \\
University of California, Berkeley \\
\texttt{moshe0110@gmail.com}
\And
Aly Lidayan \\
University of California, Berkeley
\And
Yoav Goldberg \\
Bar-Ilan University \\
Allen Institute for AI}

\newcommand{\codelink}{\footnote{Code: \url{https://github.com/Mosh0110/predictive-closure}}}
\begin{document}

\maketitle
\lhead{}
\renewcommand{\headrulewidth}{0pt}

\begin{abstract}
In joint-embedding predictive architectures (JEPA), an encoder and a predictor are trained together to produce a representation that minimizes the error in predicting the representation of the next observation.
Because the representations of both the current and the next observation are learned, it is unclear which features they will include.
Earlier characterizations describe the features JEPA selects in terms of their individual predictability, raising the concern that unpredictable features, like a traffic light, will not be encoded into the representation.
However, the prediction error is also affected by the relationships between features, e.g., the light's color helps predict whether the traffic will stop.
We propose the \textbf{predictive closure} characterization, which accounts for these relationships.
Based on a reformulation of the JEPA training objective, we show that selection depends on properties of feature \emph{sets}.
This view agrees with earlier works when features evolve independently, but makes new predictions when they do not, for instance that a feature like the traffic light can still be retained.
We test this characterization in controlled experiments, which show that unpredictable features can become better represented when they help predict other features, and that changing one feature's predictive relationships can affect how well others are represented.
We also demonstrate implications for steering feature selection: targeting one feature with an auxiliary objective, or supplying it to the predictor as side information, can strengthen the representation of another feature whose future values it helps predict, so steering can be indirect.
\end{abstract}

\section{Introduction}

We argue for a set-level characterization of how JEPA selects which features to encode.
A joint-embedding predictive architecture (JEPA) aims to learn representations useful for planning and other downstream tasks, without explicitly specifying which features those representations should include \citep{lecun2022path}.
JEPA-based systems have shown strong performance across a range of applications \citep{assran2025vjepa2,maes2026lewm,rao2026skyjepa,larey2026jepadna,zhu2026lidar,kuhn2026nova}, yet which features they encode is still far from obvious.
This matters because their representation acts as the world model JEPA builds for each observation \citep{garcia1989modelpredictive,schmidhuber1990making,sutton1991dyna,ha2018worldmodels}, so features absent from it cannot inform planning or other downstream tasks, however accurate JEPA's predictions are.

Since JEPA's encoder determines both sides of the prediction problem, JEPA learns what to predict together with how to predict it.
The encoder maps each observation (for example, three consecutive frames from a video) to a latent representation and a predictor that uses the current observation's representation to forecast the representation of the next observation.
The training objective combines a prediction loss with regularization designed to prevent the latent representation from collapsing to the same point for every input.
With a limited latent size, the encoder must leave some features out when the observation has more than fit.
Will JEPA then represent a feature whose next value is hard or impossible to predict, such as a traffic light?

\begin{figure}[!htbp]
\centering
\includegraphics[width=0.96\linewidth]{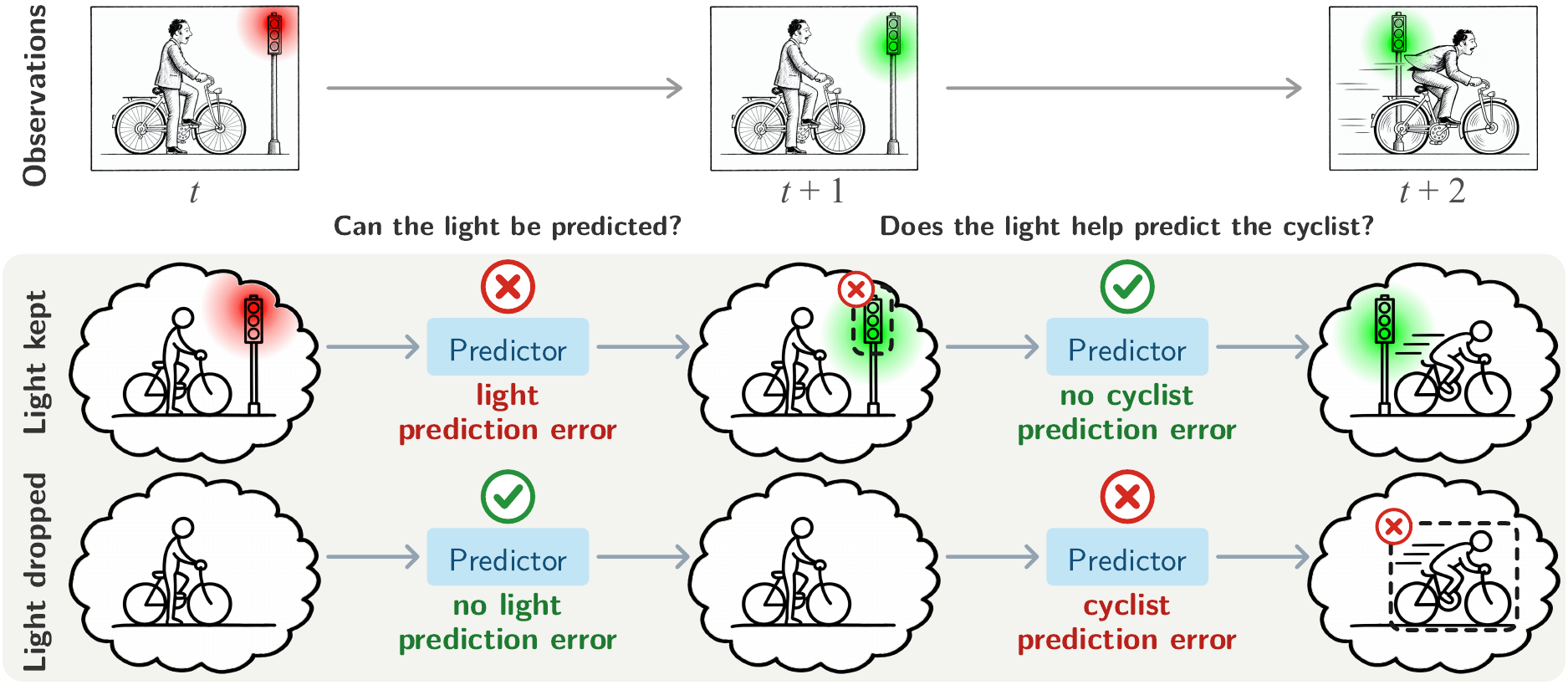}
\caption{\textbf{Will the traffic light be included in the latent representation?}
The top row shows successive observations, and the rows below show their latent representations with the light kept or dropped; crosses and dashed frames mark the resulting prediction errors.
In the first step, the light changes color at random, so keeping it costs an error on its own next color.
In the second step, the light's current color helps predict whether the cyclist will ride or stop next, so dropping the light costs an error on the cyclist.
Earlier characterizations judge the light only by whether its own next color can be predicted, and so would drop it; they disregard what its current color contributes to predicting the cyclist's next move, which is what can make the light worth keeping.
Cyclist artwork adapted from Scott McCloud's \emph{Understanding Comics} (1993).}
\label{fig:jepa-feature-selection}
\end{figure}

In recent work, \citet{pendharkar2026predictive} argues that unpredictable features such as the traffic light will not be represented.
This is consistent with the characterizations of feature selection in other works that portray it as dependent on the feature's own rate of change \citep{sobal2022slow,karmann2026motionjepapreventingtemporalfeature} or predictability \citep{littwin2024jepa,enigma2026obsessedencoder}.
All of these works judge each feature by its own behavior.
In this paper, we argue that such features can still be represented because, while hard to predict, they can help predict other features that JEPA encodes: the traffic light's current color helps predict whether a cyclist waiting at it will ride or stop next (Figure~\ref{fig:jepa-feature-selection}).
Reasoning about one feature at a time can then feel like chasing one's own tail: what is useful to represent depends on what must be predicted, while what must be predicted depends on what is represented.

To account for such relationships, we propose a novel characterization named \textbf{predictive closure}, which judges sets of features as wholes rather than each feature on its own.
It is based on a reformulation of the JEPA training objective in which selection depends on \emph{joint} properties of the selected \emph{set} of features (Section~\ref{sec:formalization}).
The reformulation shows that among the sets the latent can represent, the objective favors the set with the greatest mutual information between the current and the next values of its features.
This means that what a feature adds to JEPA's objective depends on the other features in the set.
Adding it raises the mutual information if its current value helps predict their next values, or if its next value can be predicted from their current values.
The traffic light can therefore belong to the set the objective favors, even though its own next state is impossible to predict.

Predictive closure explains what earlier works found about the features JEPA encodes, and agrees with them when features evolve independently.
When features help predict other features, it makes new predictions, including about steering methods (Section~\ref{sec:implications}).
We construct image sequences with controlled predictive relationships, train JEPA systems on them with a latent too small to hold every feature, and use probes to measure which features the latent keeps (Section~\ref{sec:experiments}).
First, we find that an unpredictable traffic-light-like feature becomes more decodable as its usefulness for predicting other features increases.
Second, strengthening one predictive connection within a set of features that help predict one another can increase its members' representation while reducing that of a competing set whose predictive relationships are unchanged.
Third, even a highly predictive set of features can remain poorly represented if it requires more capacity than the latent provides.

Predictive closure further predicts that targeting one feature during training affects the representation of its predictive partners and competitors.
We test these steering predictions in Section~\ref{sec:steering-feature-selection}.
We find that adding an auxiliary loss to reconstruct a single image can strengthen the representation of its untargeted predictive partner while reducing that of competing feature sets.
We also find that supplying a feature directly to the predictor strengthens the representation of some features while weakening others. This suggests using side information from observations to steer feature selection, even when it is not an action (Section~\ref{sec:side-information-implications}).
Both interventions can strengthen a predictive partner of the targeted feature, so steering can be indirect.
If we view actions as side information, predictive closure also suggests that action-conditioned JEPA favors the features an agent has an impact on.

\section{Background}
\label{sec:background}

A joint-embedding predictive architecture (JEPA) learns a representation by predicting the latent representation of the next observation.
Unlike reconstruction-based approaches, which train the latent to compress the information needed to recover the input \citep{kingma2014autoencoding}, JEPA does not impose this requirement, so its latent can omit some features of the observation \citep{vanassel2025joint}.
This raises the question of which features JEPA will encode, which is the main focus of this paper.
\subsection{JEPA Doesn't Explicitly Specify Features}
\label{sec:setup}

JEPA includes an encoder $f$ that maps an observation $x_t$ at time $t$ to a latent representation $f(x_t)$, and a predictor $g$ that takes $f(x_t)$ as input to predict $f(x_{t+1})$ (Figure~\ref{fig:jepa-feature-selection}).
Training combines a prediction loss on observation pairs with a regularization loss on the latent distribution \citep{lecun2022path}:
\begin{equation}
\mathcal L_{\mathrm{JEPA}}
\triangleq\mathbb E\left\|g(f(X_t))-f(X_{t+1})\right\|_2^2
+\lambda\mathcal R(f(X_t)).
\label{eq:jepa-training-objective}
\end{equation}
Here, $X_t$ denotes the random observation at time $t$, and the expectation is over consecutive observation pairs.
The parameter $\lambda>0$ controls the weight of the regularizer $\mathcal R$, which acts on the distribution of $f(X_t)$ and encourages it to match a chosen distribution to prevent collapse (such as the distributions proposed in \citealp{balestriero2025lejepa,wu2026visreg,kuang2026radial,nicollier2026spherejepa}).

Within the bounds of the enforced distribution, the objective does not specify what the latent should contain, and because the encoder produces both the predictor's input and its target, the predictor adjusts to whatever the encoder produces.

To describe what the latent represents, we use the concept of features.\footnote{An alternative way to describe the latent is to ask which differences in the observation result in the same representation, so that the latent is defined by what it omits \citep{li2006towards}.}
We define features as variables of the observation, some of which correspond to named properties such as the traffic light's color or the cyclist's speed, and we say that the latent represents a feature when the feature's value can be recovered from the latent.
See Appendix~\ref{app:formal-setup} for the formal definition that we use in our proofs.
Features allow us to treat the latent as representing separate pieces of information and to leave aside how it encodes them together.
Such a description makes it tempting to think about the latent one piece at a time and to overlook the relationships between the pieces.
It is nonetheless a common way to describe information-processing systems, both biological \citep{hubel1962receptive,marr1982vision,hung2005fast} and artificial \citep{bengio2013representation,alain2016understanding}, and it is also how earlier work has characterized the information JEPA stores, as we review next.

\subsection{Earlier Characterizations of Feature Selection}
\label{sec:prior-characterizations}

Some earlier works characterize JEPA as focusing on slow features \citep{sobal2022slow}, as encoding the most predictable features \citep{enigma2026obsessedencoder}, or as discarding unpredictable ones \citep{pendharkar2026predictive}.
Other works describe the information selected into the latent as a subspace of the observation space \citep{tang2023selfpredictive,littwin2024jepa,khetarpal2024action,vanassel2026spectral}, under two assumptions: that the encoder and predictor are linear, and that features evolve independently, so the current value of each feature carries no information about the next values of the others.

These per-feature descriptions disregard the predictive relationships \emph{between features} and assume that each feature carries information about its own future only, not about the future of other features.
In the characterization we propose in this work, we take this relation into account by assessing features jointly.
\section{Predictive Closure}
\label{sec:formalization}
\label{sec:predictive-selection}

We reformulate JEPA's prediction objective as selecting \emph{a feature set} whose present values provide the most information about \emph{its own future} (known as \emph{predictive information}) while remaining feasible under the regularization applied to the latent representation.
This yields the characterization we call \textbf{predictive closure} (Section~\ref{sec:set-selection}), which we then decompose into two terms that give a more interpretable view of the selection (Section~\ref{sec:closure-split}).

\subsection{JEPA Selects Sets, Not Individual Features}
\label{sec:set-selection}

We begin by identifying which feature sets JEPA can select.
We write $S$ for \emph{a set of features}, $Y_t(S)$ for their \emph{joint values} in an observation at time $t$, which together form one random variable, and $S(f)$ for the \emph{complete set} of features recoverable from the latent of an encoder $f$.
Our first observation is that JEPA's regularization loss constrains $Y_t(S(f))$ to a specific entropy (Section~\ref{sec:setup}), and we call a set of features $S$ \textbf{feasible} if its values have exactly this entropy (Appendix~\ref{app:formal-setup} gives the full condition).
The question is then what determines which feasible set JEPA's training objective selects.

We prove that the set $S(f^\star)$ represented by an ideally trained encoder $f^\star$, one that minimizes the training objective, is a feasible set with the greatest predictive information $I_{\mathrm{pred}}$ \citep{bialek1999predictive}:
\begin{equation}
\boxed{
S^\star\coloneqq S(f^\star)\in\operatorname*{arg\,max}_{S\ \mathrm{feasible}}
\underbrace{I\bigl(Y_t(S);Y_{t+1}(S)\bigr)}_{I_{\mathrm{pred}}(S)}.
}
\label{eq:information-predictive-closure}
\end{equation}

The optimization can be thought of as seeking a set that is most ``closed under prediction'', meaning that the current values of its features determine their next values. We call the feasible set that optimizes this the \textbf{predictive closure}.

\paragraph{Proof sketch.} With the entropy of the latent's distribution fixed, the prediction loss can be lowered only by making the next latent less uncertain given the current one, which increases the mutual information between them.
Because knowing the values of $S(f)$ is the same as knowing the latent, the mutual information between consecutive latents equals the predictive information of $S(f)$.
Maximizing it over encoders is therefore the same as maximizing predictive information over feasible sets.
See Appendix~\ref{app:prediction-information} for the full proof. 

Intuitively, adding a feature to a set of features raises the set's predictive information if the feature's value at time $t$ helps predict the values of the other features in the set at time $t+1$, or if its value at time $t+1$ can be predicted from their values at time $t$.
Here, one value helps predict another when the two have high mutual information.

The earlier characterizations in Section~\ref{sec:prior-characterizations} and predictive closure both imply that JEPA keeps features that are predictable and help predict others, and omits features that are unpredictable and help predict nothing, but they differ fundamentally in how the choice is made.
Individual selection amounts to ranking the list of features and keeping the top $n$ that fit in the latent, whereas predictive closure requires comparing alternative sets and not alternative features, because what a feature adds to the predictive information of a set depends on which other features the set contains.
The two views coincide when features evolve independently of one another: each feature then adds the same predictive information and takes up the same capacity whatever else is represented, so features compete only for capacity and ranking them is enough (see Appendix~\ref{app:ranking} for details).
Predictive closure makes new predictions when features are predictively related, in the sense that the current values of some features carry information about the value of another at the next time step.

Because features are assessed jointly, neither how easily a feature can be predicted nor how much information it provides determines by itself whether it belongs to the predictive closure.
\textbf{A feature $\phi$ will be represented if it is a member of some feasible feature set $S_{\phi}$ with greater predictive information than any other feasible set.}
These competing sets can differ in many features besides $\phi$, and some may share only constant features.
Whether one part of an observation is represented can therefore depend on whether another part is represented, even when neither helps predict the other.

\subsection{Aleatoric Uncertainty and Predictive Leakage}
\label{sec:closure-split}

We decompose the predictive information to gain further insight into the feature selection objective:
\begin{equation}
S^\star\in\operatorname*{arg\,min}_{S\ \mathrm{feasible}}
\Bigl[\underbrace{H\bigl(Y_{t+1}(S)\mid X_t\bigr)}_{\text{aleatoric uncertainty}}
+\underbrace{I\bigl(X_t;Y_{t+1}(S)\mid Y_t(S)\bigr)}_{\text{predictive leakage}}\Bigr].
\label{eq:aleatoric-leakage}
\end{equation}
The aleatoric uncertainty is how uncertain the set's next values $Y_{t+1}(S)$ remain even when the full current observation $X_t$ is known.
The predictive leakage is how well the current observation predicts the set's next values beyond what can be predicted from the set's current values, and it is zero when the set's features together predict all of their next values.
See Appendix~\ref{app:closure-split} for the derivation.

This decomposition gives us more language to describe the choice of the best set: a feature whose current value helps predict the set's next values lowers the set's predictive leakage, and one whose next value is determined by the set's current values does not add to either term.
Conversely, a noisy feature only adds to the aleatoric uncertainty unless it helps predict other features in the set, so a noisy TV that helps predict nothing \citep{burda2019largescale} will not be represented.

\section{Implications of Predictive Closure}
\label{sec:implications}

We now show how predictive closure explains earlier empirical findings on which features JEPA-based systems encode and makes predictions about what features will be selected in scenarios where features are predictively related.
Throughout this section we reason with Equation~\eqref{eq:aleatoric-leakage}: the encoder favors features that add little to its sum.

\subsection{Explaining Earlier Findings}
\citet{sobal2022slow} found that a moving object was better represented when its background was independently resampled between observations than when it remained fixed within each sequence.
A fixed background adds nothing to the sum, since its next values equal its current ones, while a resampled one adds its entire entropy as aleatoric uncertainty and is left out in favor of the moving object.
\citet{enigma2026obsessedencoder} found that even a small pattern added to the observations took over most of the latent when held fixed within each sequence, and was ignored when resampled at every observation.
Here too the resampled pattern adds its entire entropy, while every feature of the fixed pattern, however fine, adds nothing, so a set made only of such features has the smallest sum and leaves no room for the rest of the scene.

\citet{karmann2026motionjepapreventingtemporalfeature} found that a JEPA world model encoded the slowly changing score of a Pong game while suppressing the fast-moving ball, and, like \citet{sobal2022slow}, describe JEPA as biased toward slow features.
This bias follows from the time step between $t$ and $t+1$ in Equation~\eqref{eq:aleatoric-leakage}: a feature that changes little within one step is almost determined by its current value and adds almost nothing to the sum, while a fast-changing one can add far more.
What counts as slow therefore depends on the time step, so a feature can be kept at one temporal resolution and omitted at another.

\citet{pendharkar2026predictive} showed that an independently resampled visual cue, appearing over a predictably changing background, was barely decodable from JEPA's latent despite determining which action earned a reward (reward was not included in the training objective).
The cue only adds to the sum: its next value is pure aleatoric uncertainty, and its current value removes no predictive leakage.

\subsection{Predicting Feature Selection}
\label{sec:predictive-closure-implications}

\paragraph{Cross-subsidization.}

An unpredictable feature $\phi$ adds the aleatoric uncertainty of its own next value to the sum, but if $\phi$ is highly predictive of other represented features, including it also removes predictive leakage, which subsidizes this cost.
Hence, predictive closure predicts that \textbf{an unpredictable feature will be represented if it is sufficiently predictive of other represented features}.
In scenarios like the one depicted in Figure~\ref{fig:jepa-feature-selection}, where the cyclist's features are determined by the traffic light and the light itself is unpredictable, the traffic light will be represented whenever the cyclist's features are.

\paragraph{Set switching.}
Consider a scenario where the observations contain two feature sets with predictive relationships within each set, but not between them, and the latent has the capacity to fully represent either set but not both. Increasing the strength of the relationships within one set lowers the aleatoric uncertainty of its features' next values, and once its sum falls below that of the second set, it will be represented instead, even if the second set remains unchanged.
Hence, predictive closure predicts that \textbf{strengthening the predictive relationships within one set of features will eventually displace a disjoint set that competes with it for capacity}.
For example, adding sensors for an independent, highly predictable process can displace features of the original observations from the latent.\footnote{When we write of changes to the observations, we consider the counterfactual case where a JEPA system would be trained on the sequence after the change.}

\paragraph{Feasibility dependence.}
Equation~\eqref{eq:aleatoric-leakage} compares only feasible sets, and a set whose entropy exceeds the latent capacity is not feasible; each feasible part of such a set has predictive leakage through the features it leaves out.
Hence, predictive closure predicts that \textbf{enlarging a set of mutually predictive features beyond the latent capacity will reduce its representation, even if its features remain as predictable as before}, which can lead to an entirely different set being represented.

\paragraph{Steering with an auxiliary loss.}
When adding an auxiliary loss that requires some features, we force those features into the predictive closure, and the remaining features are selected such that together with the fixed features they form the set with the smallest sum.
The included features will be not only those that the auxiliary loss requires, but also their predictive partners: features whose current values help predict the required features' next values and so reduce predictive leakage, and features whose next values the required features help predict.
Hence, predictive closure predicts that \textbf{an auxiliary loss on some features will also strengthen the representation of their predictive partners, at the expense of competing features}.
For example, an auxiliary loss for reconstructing the wind measurements can favor representation of a flag's state.

\paragraph{Steering with side information.} When supplying the predictor with side information $U_t$ that it is not trained to predict (as is usually done for actions), both terms of Equation~\eqref{eq:aleatoric-leakage} are computed given $U_t$ as well.
This extends the prediction-error assumption to condition on both predictor inputs (see Appendix~\ref{app:side-information} for the derivation).
When $U_t$ is derived from the observation, only the predictive leakage changes: it becomes $I\bigl(X_t;Y_{t+1}(S)\mid Y_t(S),U_t\bigr)$, so what $U_t$ tells about the set's next values no longer leaks.
When it is not, as with actions, $U_t$ can also lower the aleatoric uncertainty.
Hence, predictive closure predicts that adding \textbf{side information can strengthen the representation of features whose next values it helps predict, without requiring the side information itself to be represented in the latent}.
For example, supplying wind measurements to the predictor can favor representation of a flag's state.

\section{Testing the Selection Predictions}
\label{sec:experiments}

We test whether trained JEPA systems follow the three selection predictions from Section~\ref{sec:predictive-closure-implications}, in image sequences that are simple by design.
Testing a prediction requires a sequence whose predictive relationships are known and can be changed one at a time, which natural data does not allow.
Each experiment therefore creates a synthetic sequence and changes a single property of it, with the architecture and the training budget held fixed, so that the change in what is represented can be attributed to that property.

\textbf{Experimental setup.}\codelink{} All experiments train JEPA on constructed observation sequences of CIFAR-10 image collages \citep{krizhevsky2009learning} and approximate whether specific \emph{target features} are represented by how accurately probes recover their values from the frozen latent on held-out observations.
These targets are the traffic-light state or binary indicators of target-image presence.
We approximate how well features of the image at each position are represented by averaging accuracy across the binary target-image detection tasks associated with that position.
We use representation of an image as shorthand for representation of its features.
We use a three-dimensional latent representation and a training objective combining the prediction loss and SIGReg regularization with coefficients of $0.9$ and $0.1$, respectively.
We choose this small latent so that even these simple sequences exceed its capacity and features must compete, as they do in larger latents on richer observations.
Unless otherwise specified, we train each configuration with three optimization seeds and report mean decoding accuracy, with error bars indicating standard errors across seeds.
Except where noted, our findings are not specific to a single choice of $\lambda$ (see sensitivity analysis in Appendix~\ref{app:regularization-sensitivity}); additional training and probing details are in Appendix~\ref{app:experimental-details}.
We validate the predictive performance of the traffic-light and two-pair trained systems in Appendix~\ref{app:prediction-validation}.

\subsection{Cross-subsidization: representing unpredictable features}
\label{sec:cross-subsidization}

\begin{figure}[!htb]
\centering
\includegraphics[width=0.88\linewidth]{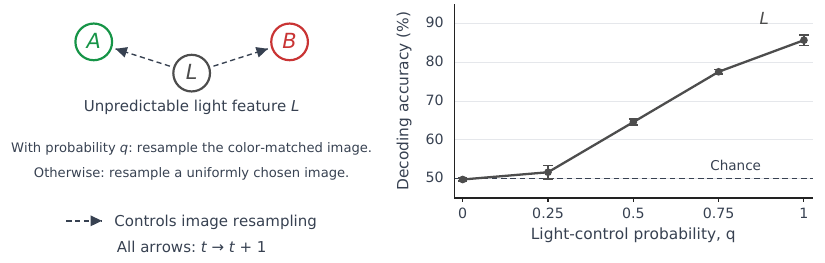}
\caption{\textbf{Cross-subsidization: the unpredictable light becomes more decodable when it helps predict the images.}
Increasing $q$ makes the current light more useful for predicting which images persist, while its future remains unpredictable.
Dashed line: $50\%$ chance accuracy.}
\label{fig:cross-subsidization}
\end{figure}

\paragraph{Scenario.}
We place a two-valued traffic light alongside two images in a collage.
We write $A$ and $B$ for the images and $L$ for the light's value, green or red.
At every transition, exactly one image is replaced by a newly sampled image while the other persists.
With probability $q$, the current light $L_t$ determines which position is resampled; otherwise, the position is chosen uniformly (Figure~\ref{fig:cross-subsidization}, left).
Green selects $A$ for replacement and red selects $B$.
The next light $L_{t+1}$ is an independent uniform draw, so the light is unpredictable and representing it adds one bit of aleatoric uncertainty at every value of $q$, while a larger $q$ lets $L_t$ remove more predictive leakage from the images' next values.

\paragraph{Results.}
As $q$ increases, the light feature $L$ becomes better represented even though its future is impossible to predict (Figure~\ref{fig:cross-subsidization}).
As predictive closure predicts, features are not selected solely according to how easy they are to predict.
Note that the transition in this scenario involves a nonlinear interaction between the light and the current images, which cannot be captured by frameworks that assume the predictor is a linear function such as \cite{littwin2024jepa}.

\subsection{Set Switching}
\label{sec:four-feature-set-tipping}

\begin{figure}[!htb]
\centering
\includegraphics[width=0.80\linewidth]{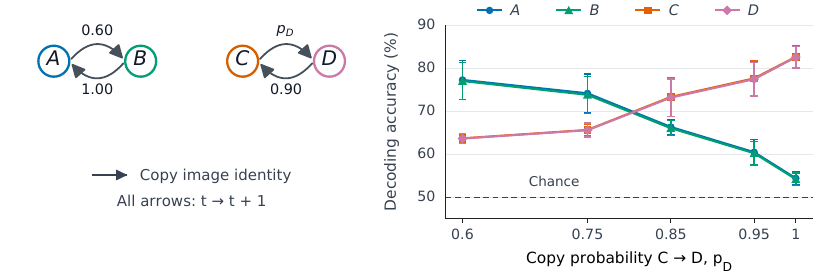}
\caption{\textbf{Set switching: changing predictive relationships reverses which pair is more strongly represented.}
Increasing $p_D$ strengthens decoding of $C/D$ and weakens $A/B$, whose predictive relationships remain fixed; neither pair helps predict the other.
Dashed line: $50\%$ chance.}
\label{fig:four-feature-set-tipping}
\end{figure}

\paragraph{Scenario.}
The two-pair setup places images $A/B$ and $C/D$ in a $2\times2$ collage.
Within each pair, an image can be copied to the other position in the next frame; otherwise, a new image is sampled independently.
We vary only the probability $p_D$ that $C_t$ is copied to $D_{t+1}$ while keeping all other copying probabilities fixed (Figure~\ref{fig:four-feature-set-tipping}, left; Appendix~\ref{app:set-setup-details}).

\paragraph{Results.}
Increasing $p_D$ shifts representation from features of $A/B$ toward features of $C/D$, although the predictive relationships within $A/B$ remain unchanged and neither pair helps predict the other (Figure~\ref{fig:four-feature-set-tipping}).
We also find similar decoding accuracy within each pair, consistent with the set-level characterization.

\subsection{Feasibility Dependence}
\label{sec:set-size-experiment}

\begin{figure}[!htb]
\centering
\includegraphics[width=0.76\linewidth]{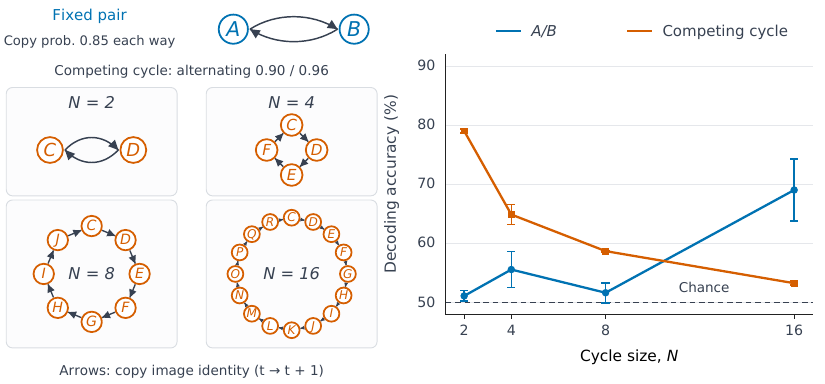}
\caption{\textbf{Feasibility dependence: adding images to a cycle can reverse which images' features are more strongly represented.}
Mean decoding accuracy across the competing cycle's positions decreases as $N$ grows, despite unchanged average copying probabilities; at $N=16$, the fixed $A/B$ pair is better represented.
Dashed line: $50\%$ chance.}
\label{fig:set-size-switching}
\end{figure}

\paragraph{Scenario.}
To test feasibility dependence, we keep $A/B$ and the latent size fixed and increase the competing cycle to $N\in\{2,4,8,16\}$ images (Figure~\ref{fig:set-size-switching}, left).
Average within-cycle copying probabilities remain unchanged, so retaining the larger cycle requires representing features of more images; tile scale and layout also change with $N$ (Appendix~\ref{app:set-setup-details}).

\paragraph{Results.}
Mean decoding accuracy for features of the competing images decreases at every increase in $N$ in all three optimization seeds (Figure~\ref{fig:set-size-switching}).
At $N=16$, features of $A/B$ are better represented in all three seeds; this endpoint reversal also holds for the worst and best positions in each cycle (Appendix~\ref{app:capacity-position-readouts}).
In one out of the three sensitivity-analysis configurations, the increase in representation of $A/B$'s features is within the error bars (see Appendix~\ref{app:regularization-sensitivity}).

    \section{Testing the Steering Predictions}
\label{sec:steering-feature-selection}

We now test the two steering predictions from Section~\ref{sec:predictive-closure-implications}.
Since selection operates jointly, we examine how steering that targets features of a single image affects both their predictive partners and competing features.
Both experiments use the two-pair setup from Section~\ref{sec:four-feature-set-tipping}, keeping the sequence's predictive relationships, image pool, three-dimensional main latent representation, and training budget unchanged.
We measure image-detection accuracy from the frozen main encoder alone, training probes on collages of images held out from encoder training and evaluating them on separate collages.
Chance decoding accuracy is $50\%$ for each feature individually.

\subsection{An Auxiliary Reconstruction Objective}
\label{sec:auxiliary-objective}

Auxiliary objectives have been proposed to steer JEPA representations toward task-relevant information \citep{lecun2022path,destrade2025value,zhu2026spectral,vujinovic2026actjepa}.
An auxiliary reconstruction objective encourages the encoder to retain features of the image being reconstructed.
Section~\ref{sec:predictive-closure-implications} predicts that this incentive will also favor their predictive partners, at the expense of competing features.

\paragraph{Steering.}
To test this prediction, we add an auxiliary objective that reconstructs a coarse version of the current $C$ image from the main latent representation; only $C$ is a reconstruction target, and the decoder is discarded before probing.
We vary the reconstruction weight $\beta$ across all five original values of $p_D$, using one optimization seed.
The nonzero weights come from a preliminary search with both $C$ and $D$ as reconstruction targets (Appendix~\ref{app:steering-details}).

\begin{figure}[!htb]
\centering
\includegraphics[width=\linewidth]{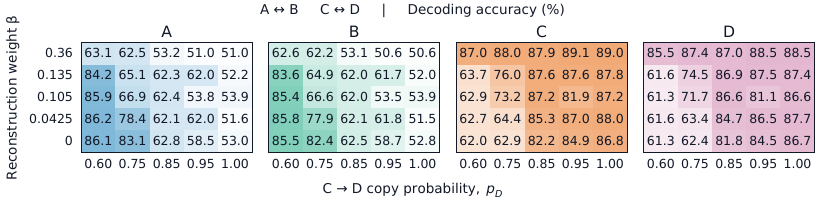}
\caption{\textbf{Reconstructing one image can strengthen representation of its predictive partner.}
The changes are not uniformly monotonic, and the baseline already favors $C/D$ at high $p_D$.
Each cell uses one optimization seed; all panels use the same $50$--$90\%$ color scale.}
\label{fig:auxiliary-reconstruction}
\end{figure}

\paragraph{Results.}
Rewarding reconstruction of $C$ alone can make $D$'s features more strongly represented and $A/B$'s less so (Figure~\ref{fig:auxiliary-reconstruction}).
At the largest weight, $D$, which is never reconstructed, is decoded almost as accurately as $C$ at every $p_D$.
This supports the prediction that steering can affect representation beyond the directly targeted features.

\subsection{Predictor Side Information}
\label{sec:predictor-side-information}

Supplying information directly to the predictor changes how well it can predict different features.
Section~\ref{sec:predictive-closure-implications} predicts that this can strengthen the representation of features whose next values the additional information helps predict.

\paragraph{Steering.}
To test this prediction, we supply the current $C$ image through an additional predictor input, without an auxiliary loss or masking it from the main input.
A separate encoder with the same architecture as the main encoder maps $C$ to a three-dimensional vector, which we concatenate with the main latent representation before prediction; this adds encoder capacity while leaving the main representation's dimension unchanged.
The side encoder is trained jointly with the main encoder and predictor through the prediction loss; SIGReg is applied only to the main latent representation.
We evaluate all five values of $p_D$, paired with their no-side-input controls.
We validate these systems' predictive performance in Appendix~\ref{app:prediction-validation}.

\begin{figure}[!htb]
\centering
\includegraphics[width=\linewidth]{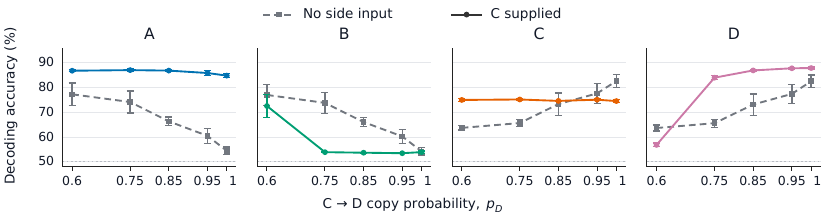}
\caption{\textbf{Supplying $C$ to the predictor can strengthen representation of its predictive partner $D$.}
Data and probe tasks are matched; probes use only the main latent.
$D$ becomes more represented for $p_D\geq0.75$ and less at $p_D=0.60$, where $C$ is least often copied to $D$; $C$ becomes less represented at the highest $p_D$.
Dotted line: $50\%$ chance.}
\label{fig:predictor-side-information}
\end{figure}

\paragraph{Results.}
Supplying $C$ changes how well features of the other images are represented, but produces a different redistribution from reconstruction (Figure~\ref{fig:predictor-side-information}).
For most values of $p_D$, we see an increase in the representation of $D$'s features and a decrease in that of $B$'s, while representation of $A$'s features increases throughout the sweep.
These changes support the steering prediction for $D$; predictive closure does not account for $A$ and $B$ separating, since it values them only as a pair.
Low decoding accuracy from the main latent alone does not establish that a feature is unavailable to the predictor: information about $C$ or other features may be recoverable jointly from the main and side representations.

\subsection{Implications for representation learning}
\label{sec:side-information-implications}

\paragraph{Action conditioning.}
In action-conditioned JEPA, the predictor receives the agent's action alongside the current latent.
If we view the side information as an action, features that the agent \emph{has an impact on} are more likely to be represented.
The action helps predict their future values, so the sets that include them gain the predictive information induced by their relationships with the action (Appendix~\ref{app:side-information}).
Features that the agent does not affect gain no such information and are less likely to be represented, unless they are predictive partners of features the agent affects.
In our side-information experiment, $C$ plays the role of the action and $D$ of a feature it affects: for $p_D=0.75,0.85,0.95$, supplying $C$ strengthens representation of $D$.
The consequences of genuine action conditioning remain to be tested.

\paragraph{Steering through extracted information.}
A practitioner can instead choose side information specifically to steer predictive closure.
If a desired set of predictive features is poorly represented and one of its features can be extracted from the observation, supplying that feature to the predictor may strengthen representation of the features whose next values it helps predict.
Our side-information experiment illustrates this possibility: supplying $C$ can strengthen representation of $D$ without directly supervising $D$.
The mixed changes across the remaining features show why steering should be evaluated against all features needed downstream.

\paragraph{Steering can be indirect.}
Both forms of steering target only $C$, yet both can strengthen representation of $D$.
This shows that steering can be indirect: it need not target the desired feature itself.
Indirect steering is particularly useful when a related feature can be targeted but no loss can be written for the desired feature.
An auxiliary loss can target a feature that helps predict its next values, or one whose next values it helps predict; side information can supply only a feature of the first kind.
For example, an object's velocity determines where the object will be next, so an auxiliary loss on its velocity, or supplying the velocity to the predictor, can strengthen the representation of its position.

\section{Conclusion}
\label{sec:conclusion}

We characterize JEPA representation learning through the joint predictive value of feature sets.
We reformulate the training objective as selecting a \textbf{predictive closure}: a feasible feature set that maximizes the mutual information between its features' current values and their next values.
This view explains earlier findings on other JEPA systems, and our experiments match its predictions: an unpredictable feature becomes more decodable when it helps predict others, and changing a competing set can weaken a feature's representation even when its own predictive relationships are unchanged.
Experiments with auxiliary objectives and predictor side information further show that targeting one feature can strengthen the representation of its predictive partners, so steering can be indirect.
If we view actions as side information, predictive closure also suggests that action-conditioned JEPA favors the features an agent has an impact on.
These experiments illustrate how this view can inform the design of JEPA systems to better represent task-relevant features.
Our experiments test the direction of each predicted effect; measuring its size in large JEPA systems trained on natural data remains for future work.
This set-level view may also be relevant to models of biological representation learning that link representation to prediction \citep{palmer2015predictive,chalk2018unified,singer2023hierarchical}.
We hope predictive closure provides a clearer way to reason about which features JEPA encodes and informs the design and use of these systems in downstream tasks.

\bibliographystyle{iclr2027_conference}
\bibliography{references}

\clearpage
\appendix
\section{Formalization of Predictive Closure}
\label{app:prediction-information}

We provide a supporting derivation for the predictive-closure characterization in Equation~\eqref{eq:information-predictive-closure} of Section~\ref{sec:predictive-selection}.
We first give the definitions and assumptions, then connect prediction loss to mutual information and express the resulting objective in terms of feature sets.

\subsection{Definitions and Assumptions}
\label{app:formal-setup}

We use features to describe observations through identifiable properties, such as the traffic light's color or whether the cyclist is riding or stopping.
This lets us characterize what JEPA encodes in terms of those properties and their predictive relationships.
The latent could also be described by which observations it maps to the same representation, so that the latent is defined by what it omits: whatever differs among observations that share a representation \citep{givan2003equivalence,li2006towards}.

\label{sec:feature-definitions}
\paragraph{Features.}
A feature $\phi_i$ describes a property of an observation, such as the traffic light's color or the cyclist's speed.
Formally, $\phi_i$ is a mapping from observations to scalar values, and we write $y_t^i=\phi_i(x_t)$ for its value at time $t$.
This excludes properties that cannot be inferred from the observation, such as which watch the cyclist wears when it is completely hidden and no other clues reveal it.
Different mappings define different features.
For example, a mapping that matches whether the cyclist is riding or stopping on some observations but not others defines a different feature.
A \textbf{feature set} $S=\{\phi_1,\ldots,\phi_m\}$ is a collection of these mappings.
We write $Y_t(S)=(\phi_i(X_t))_{\phi_i\in S}$ for their values on an observation $X_t$ at time $t$.
Likewise, a \textbf{reader} $d$ is a mapping from latents to scalar values, like those the predictor or a downstream model uses to read values off the latent.

\paragraph{Represented feature sets.}
We say that a feature $\phi_i$ is \textbf{represented} in the latent $f(x_t)$ when a reader $d_i$ whose output changes only slightly under small changes to the latent (formalized in Appendix~\ref{app:feature-selection}) recovers its value across observations: $d_i(f(x_t))=y_t^i$ for all $t$.
We write $S(f)$ for the complete set of features recoverable in this way from the latent produced by encoder $f$.

\label{sec:theory-setup}

\paragraph{Fixed latent entropy.}
Representing a feature requires capacity \citep{klindt2026unifying} and we assume that all encoders match the same Gaussian target prescribed by regularization at each time, with $P$ denoting a finite-precision latent distribution.
\footnote{All encoders use the same numerical precision, and readers access the stored values. This idealizes the distribution matching encouraged by regularizers such as SIGReg, for which \citet[Section~4.3]{balestriero2025lejepa} provide empirical support.}
The shared distribution fixes the Shannon entropy of the latent distribution $H(P)$, which we refer to as the latent's \textbf{entropy constraint}.
We assume that the entropy constraint $H(P)$ is less than the observations' entropy so not all features can be represented. We call a feature set $S$ \textbf{feasible} when some encoder represents all of its features and $H(Y_t(S))=H(P)$.
The complete set $S(f)$ of any encoder $f$ is feasible, since its values and the latent determine one another, so that $H(Y_t(S(f)))=H(f(X_t))=H(P)$ (Appendix~\ref{app:feature-selection}).
If we could directly choose which features are represented, adding a feature that provides new information would require removing others to keep the resulting set feasible.

Throughout, the prescribed distribution $P$ and latent dimension $k$ are fixed across encoders.

\subsection{Prediction Loss and Information}
\label{app:gaussian-prediction-information}

Because the regularization term in Equation~\eqref{eq:jepa-training-objective} depends only on the distribution of $f(X_t)$, which is $P$ for every encoder, an encoder minimizes the training objective exactly when it minimizes the prediction loss.
Under a Gaussian prediction-error assumption, we show that minimizing expected squared prediction error is equivalent to maximizing the mutual information between consecutive representations.
We write $f^\star$ for any such encoder:
\begin{equation}
f^\star\in\operatorname*{arg\,min}_{f}\;
\min_g\,\mathbb E\left\|g(f(X_t))-f(X_{t+1})\right\|_2^2
=\operatorname*{arg\,max}_{f}\;
I\bigl(f(X_t);f(X_{t+1})\bigr).
\label{eq:temporal-information-objective}
\end{equation}
The expectation is over consecutive observation pairs.

\paragraph{Prediction-error assumption.}
Write $Z_t=f(X_t)$.
We assume that all encoders produce $k$-dimensional latents with the same marginal distribution $P$, whose Shannon entropy $H(P)$ is finite.
For each encoder $f$, assume that among the functions the predictor network can represent there is a predictor $g_f$ such that, at the latent's numerical resolution,
\begin{equation}
Z_{t+1}\mid Z_t=z
\sim\mathcal N\!\left(g_f(z),\sigma_f I_k\right),
\label{eq:gaussian-prediction-error}
\end{equation}
where $\sigma_f>0$ is a scalar variance that may differ between encoders.

\paragraph{Proof.}
For each current latent, squared prediction error is minimized by the mean of this Gaussian, which $g_f$ predicts.
Each error coordinate has zero mean and variance $\sigma_f$, so its expected squared error is $\sigma_f$.
Summing over the $k$ coordinates gives
\begin{equation}
\min_g\mathbb E\left\|Z_{t+1}-g(Z_t)\right\|_2^2
=k\sigma_f.
\label{eq:gaussian-optimized-mse}
\end{equation}
The conditional Gaussian distribution has entropy
\begin{equation}
h(Z_{t+1}\mid Z_t)
=\frac{k}{2}\log(2\pi e\sigma_f).
\label{eq:gaussian-prediction-conditional-entropy}
\end{equation}
At the latent's fixed numerical precision, we assume the discrete conditional entropy equals this Gaussian value plus an additive correction $c$ shared across encoders.
($c$ comes from rounding the continuous latent to that precision, so it depends only on the precision.)
Since the next latent has the fixed marginal distribution $P$,
\begin{equation}
\begin{aligned}
I(Z_t;Z_{t+1})
&=H(P)-H(Z_{t+1}\mid Z_t)\\
&=H(P)-\frac{k}{2}\log(2\pi e\sigma_f)-c.
\end{aligned}
\label{eq:gaussian-mse-information}
\end{equation}
Because $H(P)$, $k$, and $c$ are fixed, predictive information is strictly decreasing in $\sigma_f$, and so in the optimized prediction error $k\sigma_f$.
Therefore, minimizing expected squared prediction error and maximizing predictive information select exactly the same encoders.
Without the Gaussian assumption, and with the same precision correction $c$, Equation~\eqref{eq:gaussian-mse-information} holds with $\geq$ in place of its second equality for every encoder that matches $P$, with $\sigma_f$ taken as its optimized prediction error divided by $k$, because among distributions with a given summed variance the isotropic Gaussian has the largest entropy (averaging over $Z_t$ uses the concavity of the logarithm).
Minimizing the prediction loss then maximizes a lower bound on the mutual information between consecutive latents.

\subsection{From Representations to Feature Sets}
\label{app:feature-selection}

We now write the objective of Equation~\eqref{eq:temporal-information-objective} in terms of the complete set $S(f)$.

\paragraph{Reader condition.}
For $Z=f(X_t)\sim P$, we control reader sensitivity using common tolerances $\varepsilon,\delta>0$ and $0<\alpha<1$, keeping these tolerances unchanged across encoders:
\[
\Pr\!\left[
\sup_{\|z'-Z\|_2\leq\varepsilon}|d(z')-d(Z)|>\delta
\right]\leq\alpha.
\]
Thus, latent perturbations of size at most $\varepsilon$ change the reader's output by at most $\delta$, except on a set of probability at most $\alpha$.
We require exact recovery from the unperturbed latent.

\paragraph{Information preservation.}
The latent determines every feature value in $Y_t(S(f))$.
Conversely, the readers $d_j(z)=(\delta/\varepsilon)z_j$, for $j=1,\ldots,k$, satisfy the sensitivity condition everywhere, so the corresponding features $x\mapsto d_j(f(x))$ belong to $S(f)$.
Their values recover every latent coordinate.
The latent and the values $Y_t(S(f))$ therefore determine one another, so they carry the same information, and their mutual information across consecutive observations is equal:
\begin{equation}
I\bigl(f(X_t);f(X_{t+1})\bigr)
=I\bigl(Y_t(S(f));Y_{t+1}(S(f))\bigr).
\label{eq:feature-information-equivalence}
\end{equation}
The same equality holds for every feasible set $S$, with $f$ an encoder that represents it: the latent determines the values $Y_t(S)$, and a function of the latent that keeps its entropy $H(P)$ is one-to-one, so these values also determine the latent.
This equality and Appendix~\ref{app:gaussian-prediction-information} together show that $S(f^\star)$ maximizes predictive information over feasible feature sets, proving Equation~\eqref{eq:information-predictive-closure}.

\subsection{Aleatoric Uncertainty and Predictive Leakage}
\label{app:closure-split}

For a feature set $S$, we write $Y_t=Y_t(S)$ and $Y_{t+1}=Y_{t+1}(S)$.
By definition,
\[
I(Y_t;Y_{t+1})=H(Y_{t+1})-H(Y_{t+1}\mid Y_t),
\]
and for a feasible set the first term is $H(P)$.
Conditional mutual information is defined in the same way, given $Y_t$:
\[
I(X_t;Y_{t+1}\mid Y_t)=H(Y_{t+1}\mid Y_t)-H(Y_{t+1}\mid X_t,Y_t).
\]
Because $Y_t$ is computed from $X_t$, knowing $X_t$ already means knowing $Y_t$, so $H(Y_{t+1}\mid X_t,Y_t)=H(Y_{t+1}\mid X_t)$ and
\[
H(Y_{t+1}\mid Y_t)=H(Y_{t+1}\mid X_t)+I(X_t;Y_{t+1}\mid Y_t).
\]
Substituting this into the first equation gives
\begin{equation}
I(Y_t;Y_{t+1})=H(P)-H(Y_{t+1}\mid X_t)-I(X_t;Y_{t+1}\mid Y_t),
\label{eq:closure-split}
\end{equation}
so among feasible sets, maximizing predictive information is the same as minimizing the sum of the last two terms, as in Equation~\eqref{eq:aleatoric-leakage}.

\subsection{Relation to Per-Feature Selection}
\label{app:ranking}

For a feature set $S$ and a feature $\phi\notin S$, we write $Y_t=Y_t(S)$ and $\phi_t=\phi(X_t)$.
By the chain rule of mutual information, adding $\phi$ to $S$ changes the set's predictive information by
\begin{equation}
\Delta(\phi\mid S)\coloneqq I_{\mathrm{pred}}(S\cup\{\phi\})-I_{\mathrm{pred}}(S)
=I\bigl(\phi_t;Y_{t+1}\mid Y_t\bigr)+I\bigl((Y_t,\phi_t);\phi_{t+1}\mid Y_{t+1}\bigr),
\label{eq:marginal-contribution}
\end{equation}
and changes its entropy by $H(\phi_t\mid Y_t)$, the \textbf{entropy cost} of $\phi$.
The first term is the information that $\phi$'s current value adds about the set's future.
The second is the information that the set's present, together with $\phi$'s, provides about $\phi$'s future beyond the set's own future.
Both terms and the entropy cost generally depend on $S$.
In optimization terminology, predictive information is generally a non-additive, or non-modular, set objective: its value depends on the combination of selected features rather than a sum of fixed individual contributions.

\paragraph{Independent features.}
Suppose features $\phi_1,\ldots,\phi_n$ evolve independently, so that the pairs $(\phi_i(X_t),\phi_i(X_{t+1}))$ are mutually independent across $i$.
For a set $S$ of these features and $\phi_j\notin S$, the first term of Equation~\eqref{eq:marginal-contribution} vanishes, the second equals $I(\phi_j(X_t);\phi_j(X_{t+1}))$, and the entropy cost equals $H(\phi_j(X_t))$, none of which depends on $S$.
Hence
\[
I_{\mathrm{pred}}(S)=\sum_{\phi_i\in S}I\bigl(\phi_i(X_t);\phi_i(X_{t+1})\bigr),
\qquad
H\bigl(Y_t(S)\bigr)=\sum_{\phi_i\in S}H\bigl(\phi_i(X_t)\bigr).
\]
If these features all have the same entropy and $H(P)$ is $k$ times it, then every feasible set formed from whole features $\phi_i$ contains $k$ of them, and the one with the greatest predictive information consists of the $k$ features with the greatest individual predictive information $I(\phi_i(X_t);\phi_i(X_{t+1}))$.
Selection among such sets then reduces to ranking individual features.
With unequal entropies, it becomes a knapsack problem in which each feature's value and cost are still independent of the other features selected.

\paragraph{Dependent features.}
For the traffic light $L$ of Section~\ref{sec:cross-subsidization} and a set $S$ of image features, $L_{t+1}$ is independent of $(Y_t,L_t,Y_{t+1})$, so $\Delta(L\mid S)=I(L_t;Y_{t+1}\mid Y_t)$.
This contribution is zero unless $S$ contains the images whose persistence the light controls, while its entropy cost $H(L_t\mid Y_t)$ is one bit at every $q$.
In the cycle of Section~\ref{sec:set-size-experiment}, the information an image carries through its copy link counts only when the image it is copied to is also in the set.
A set holding part of a long cycle therefore keeps only the links whose two ends it contains.

\subsection{Side Information}
\label{app:side-information}

Section~\ref{sec:predictive-closure-implications} considers a predictor that also receives side information $U_t$, so that it predicts $f(X_{t+1})$ by $g(f(X_t),U_t)$.
We derive how this changes the two terms of Equation~\eqref{eq:aleatoric-leakage}.
For a feature set $S$, we again write $Y_t=Y_t(S)$ and $Y_{t+1}=Y_{t+1}(S)$.

\paragraph{Objective.}
We extend the prediction-error assumption of Appendix~\ref{app:gaussian-prediction-information} to condition on both predictor inputs: writing $Z_t=f(X_t)$, for each encoder $f$ there is a predictor $g_f$ such that
\[
Z_{t+1}\mid Z_t=z,\,U_t=u
\sim\mathcal N\!\left(g_f(z,u),\sigma_f I_k\right).
\]
The proof in Appendix~\ref{app:gaussian-prediction-information} then holds with $U_t$ added to every condition, so the optimized prediction error is strictly increasing in $H(Z_{t+1}\mid Z_t,U_t)$.
Because the latent and the values of the complete represented set determine one another (Appendix~\ref{app:feature-selection}), this entropy equals $H(Y_{t+1}\mid Y_t,U_t)$ for $S=S(f)$.
An ideally trained encoder therefore represents a feasible set that minimizes $H(Y_{t+1}\mid Y_t,U_t)$, which is the same as maximizing $I\bigl((Y_t,U_t);Y_{t+1}\bigr)=H(P)-H(Y_{t+1}\mid Y_t,U_t)$.

\paragraph{The two terms.}
The steps of Appendix~\ref{app:closure-split} hold with $U_t$ added to every condition.
By the definition of conditional mutual information,
\[
I(X_t;Y_{t+1}\mid Y_t,U_t)=H(Y_{t+1}\mid Y_t,U_t)-H(Y_{t+1}\mid X_t,Y_t,U_t),
\]
and because $Y_t$ is computed from $X_t$, the last term equals $H(Y_{t+1}\mid X_t,U_t)$, so
\begin{equation}
H(Y_{t+1}\mid Y_t,U_t)=H(Y_{t+1}\mid X_t,U_t)+I(X_t;Y_{t+1}\mid Y_t,U_t).
\label{eq:side-information-split}
\end{equation}
The minimized quantity is thus the sum of the two terms of Equation~\eqref{eq:aleatoric-leakage}, each computed given $U_t$ as well.

\paragraph{Change in each term.}
By the definition of conditional mutual information, supplying $U_t$ lowers the sum by
\[
H(Y_{t+1}\mid Y_t)-H(Y_{t+1}\mid Y_t,U_t)=I(U_t;Y_{t+1}\mid Y_t),
\]
which is what $U_t$ tells about the set's next values beyond the set's current values, so sets for which this quantity is larger are favored.
Supplying $U_t$ lowers the aleatoric uncertainty by
\[
H(Y_{t+1}\mid X_t)-H(Y_{t+1}\mid X_t,U_t)=I(U_t;Y_{t+1}\mid X_t),
\]
which is what $U_t$ tells about the set's next values beyond the whole current observation.
When $U_t$ is derived from the observation, knowing $X_t$ already means knowing $U_t$, so this second quantity is zero.
The aleatoric uncertainty is then unchanged, and the whole reduction comes from the predictive leakage:
\[
I(X_t;Y_{t+1}\mid Y_t,U_t)=I(X_t;Y_{t+1}\mid Y_t)-I(U_t;Y_{t+1}\mid Y_t).
\]
When $U_t$ is not determined by the observation, as with an action, the second quantity can be positive.
For example, an agent's next position can be uncertain given the whole current observation and certain once its action is known.

\section{Experimental Setup and Evaluation Details}
\label{app:experimental-details}

\subsection{Shared training and probing protocol}

In each experiment in Section~\ref{sec:experiments}, we vary a single design parameter while keeping the remaining data-generation rules fixed.
At each value of the varied parameter, we train the same end-to-end JEPA from scratch on sequences of collages composed of CIFAR-10 images, using a three-dimensional latent and the objective $0.9\mathcal{L}_{\mathrm{pred}}+0.1\mathcal{L}_{\mathrm{SIGReg}}$ \citep{balestriero2025lejepa}.
Training draws from a class-balanced set of 5{,}000 images from the official training split; after training, we freeze the encoder and train probes to decode designated features from newly sampled collages drawn from a disjoint class-balanced set of 5{,}000 test images.

\paragraph{Architecture and gradients.}
The encoder has four $3\times3$ convolution/ReLU/$2\times2$ max-pooling blocks with 32, 64, 128, and 256 channels, followed by a $4096\to1024\to3$ projection with a GELU hidden activation.
The group-size encoder additionally uses adaptive max-pooling to produce a $4\times4$ spatial output.
Predictors use $3\to16\to16\to3$ GELU MLPs, with six inputs for side information; the reconstruction decoder is an affine $3\to48$ map.
We share the main encoder across frames and backpropagate through both prediction-loss branches, without stop-gradient.
All JEPA systems train with AdamW at learning rate $10^{-3}$ with cosine decay to $10^{-5}$, and SIGReg uses 1{,}024 random projections.

\subsection{Two-image traffic-light setup and readouts}
\label{app:traffic-light-details}

To vary the light's usefulness for predicting the next transition while keeping each image's marginal persistence fixed, we let the light control which image is replaced.
The two $32\times32$-pixel images occupy the upper row of a $64\times64$-pixel canvas, and a $4\times4$-pixel light patch appears in the lower-right quadrant; the remaining pixels are black.
Each transition replaces exactly one image with a different identity; the current light selects the position with probability $q$, and a uniform draw selects it otherwise.
The light is sampled independently and uniformly at every frame.
Consequently, each image persists with marginal probability $0.5$ at every $q$, while the light identifies the replaced position with probability $(1+q)/2$.
We train three runs at each $q\in\{0,0.25,0.5,0.75,1\}$, each for 25 epochs of 100{,}000 eight-frame sequences with batch size 512.

A two-hidden-layer probe of width 16 decodes the current light from the frozen current-frame latent, using separately sampled training, validation, and test datasets containing 20{,}000, 5{,}000, and 5{,}000 examples from the held-out image pool, respectively.
The probe trains for 1{,}200 steps with batch size 512; chance accuracy is $50\%$.

\subsection{Setups and readouts for the two-pair and cycle experiments}
\label{app:set-setup-details}

\paragraph{Changing predictability.}
The two-pair setup groups image identities into $\{A,B\}$ and $\{C,D\}$ at the four locations of a $2\times2$ collage.
Within each pair, an image identity can be copied to the other position in the next frame: $B_t$ determines $A_{t+1}$ with probability $1$, $A_t$ determines $B_{t+1}$ with probability $0.6$, and $D_t$ determines $C_{t+1}$ with probability $0.9$.
The probability $p_D$ that $C_t$ determines $D_{t+1}$ is the only parameter we vary; when a copy does not occur, the next image is sampled independently.
We train three runs at each $p_D\in\{0.60,0.75,0.85,0.95,1\}$, each for 25 epochs of 100{,}000 eight-frame sequences with batch size 512.

\paragraph{Changing group size.}
The $A/B$ cycle is fixed at two identities with copy probability $0.85$, and the competing $C/D$ cycle has $N\in\{2,4,8,16\}$ identities, ordered $C,D,E,F,\ldots$ through the first $N$ positions, with copy probabilities alternating between $0.90$ and $0.96$.
Thus the average per-feature copy probabilities remain $0.85$ for $A/B$ and $0.93$ for $C/D$ at every size.
We train three runs per configuration, each for 100 epochs of 100{,}000 eight-frame sequences, with batch size 512 and a three-dimensional latent.
Within each design, $\{A,B\}$ remains unchanged, allowing us to assess whether changes in the competing group alter its representation.
We used the first run during configuration development and kept the configuration unchanged for the two replications; all three runs share the image pools and data-generation randomness.

\paragraph{Rendering control.}
Keeping individual image size fixed would make the larger $C/D$ group occupy more of the input, confounding feature count with the group's total spatial extent.
We therefore keep the $128\times256$-pixel canvas and the two native $32\times32$-pixel $A/B$ images fixed, while distributing approximately the same total displayed area among the $C/D$ images.
To avoid removing image detail through downsampling, we use the native CIFAR-10 resolution as the minimum tile size and enlarge images when fewer positions share this area.
For $N=2,4,8,16$, individual $C/D$ images have side lengths of 90, 64, 45, and 32 pixels, respectively, giving occupied areas of 16{,}200, 16{,}384, 16{,}200, and 16{,}384 pixels, which differ by less than $1.2\%$.
The $C/D$ images occupy centered row-major grids of $1\times2$, $2\times2$, $2\times4$, and $4\times4$ within the same right-hand region; unused pixels remain black.
Larger tiles use bilinear enlargement of the native CIFAR-10 images, without cropping or reducing any image below its original resolution.
Enlargement is greatest at $N=2$ and goes back to the original size at $N=16$.
Tile scale and layout still change with $N$, so this control does not isolate representation cost from the difficulty of extracting the images' information.

\paragraph{Image-detection probes.}
To assess image-identity information in the latent, we train target-image presence detectors separately for each fixed position.
The group-size experiment uses the same primary image-detection probe architecture, per-detector training budget, and held-out image pool as the four-image predictability and steering experiments.
We train a separate binary classifier for each of 50 target images at every fixed position, with five targets per CIFAR-10 class.
Each classifier receives only the frozen latent representation: positive examples contain its target at the designated position, while negative examples exclude that target everywhere and use a different image of the same class at the designated position.
Other positions are sampled independently, excluding the target.
Because negative examples exclude the target everywhere, a representation of the unordered images within a pair could pass both position-associated probes without distinguishing their positions.
These readouts measure target-image detection under the stated sampling protocol and do not establish position binding.

Each detector has two hidden layers of width 16 and trains for 1{,}000 steps with batch size 128 on 512 training examples, with 512 validation and 1{,}024 test examples, balanced between positive and negative cases.
We use AdamW with learning rate $10^{-3}$ decaying to $10^{-5}$, 50 warmup steps, weight decay $10^{-4}$, gradient clipping at 10, and the same probe optimization seed across comparisons.
Latent coordinates are standardized using 100{,}000 separately sampled collages from the held-out image pool.
All probe images are unseen during encoder training; target identities recur across probe training, validation, and testing, while the collages are sampled separately from the same held-out pool.
We report ordinary binary classification accuracy at the zero-logit threshold, with chance accuracy $50\%$.
Increasing the number of positions increases the number of independent detectors, rather than reducing the training or evaluation budget per position.
We first average accuracy over target images within each position, then average positions within each group, and finally report the mean and standard error across optimization seeds.

\subsection{Steering interventions}
\label{app:steering-details}

Both interventions use the two-pair setup, keeping the sequence's predictive relationships, image pool, three-dimensional main latent representation, and training budget unchanged.
Held-out image-detection probes receive only the frozen main encoder's representation.

\paragraph{Auxiliary reconstruction.}
To directly reward retaining information about $C$, we add an auxiliary reconstruction objective.
The decoder reconstructs a standardized $4\times4$ average-pooled version of the current $C$ image from the current latent representation, giving the objective $0.9\mathcal{L}_{\mathrm{pred}}+0.1\mathcal{L}_{\mathrm{SIGReg}}+\beta\mathcal{L}_{\mathrm{recon}}^{C}$.
Only $C$ is a reconstruction target; the decoder is discarded before probing.
The four nonzero weights were identified in a preliminary search at $p_D=0.75$ with both $C$ and $D$ as reconstruction targets.
We then evaluate reconstruction of $C$ alone at $\beta\in\{0,\allowbreak 0.0425,\allowbreak 0.105,\allowbreak 0.135,\allowbreak 0.36\}$ and all five original values of $p_D$, using one optimization seed.

\paragraph{Predictor side information.}
To examine how direct predictor access to $C$ changes the main representation, we supply the current $C$ image through a separate encoder.
The side encoder has the same architecture as the main encoder and maps the current $32\times32$-pixel $C$ image, bilinearly enlarged to $64\times64$ pixels, to a three-dimensional vector.
We concatenate this vector with the current main latent to form the predictor's six-dimensional input; its target remains the next-frame main latent.
The main encoder continues to receive the full collage, including $C$, and retains its three-dimensional output.
Training uses $0.9\mathcal{L}_{\mathrm{pred}}+0.1\mathcal{L}_{\mathrm{SIGReg}}$, with the side encoder trained through the prediction loss and SIGReg applied only to the main latent.
This intervention adds encoder capacity while keeping the main latent dimension fixed.
We evaluate all five values of $p_D$ with three training runs per condition, paired with their corresponding no-side-input controls.

\section{Next-Observation Prediction}
\label{app:prediction-validation}

To validate that the JEPA systems learned to predict, we test whether their predictions distinguish the actual next observation from an alternative.
We evaluate the traffic-light systems trained with light-control probability $q=1$, in which the current light always determines which image is replaced (Section~\ref{sec:cross-subsidization}).
We also evaluate the two-pair systems at all five values of $p_D$, the probability that $C_t$ is copied to $D_{t+1}$ (Section~\ref{sec:four-feature-set-tipping}).
For these, we test both the ordinary systems and those whose predictor also receives the current $C$ image as side information (Section~\ref{sec:predictor-side-information}).
Training settings follow Appendix~\ref{app:experimental-details}.

We evaluate each trained system on $5{,}000$ transitions generated from images held out from encoder training, using the same transitions across runs.
For each transition, the system predicts the latent of the next observation from the current observation $x_t$.
We count the prediction as correct if it is closer, in squared Euclidean distance, to the latent $f(x_{t+1})$ of the actual next observation than to the latent $f(\tilde x)$ of an alternative observation $\tilde x$.
We draw $\tilde x$ from the same distribution as the observations and redraw it whenever the transition rules allow it to follow $x_t$.
It is therefore as likely as $x_{t+1}$ overall and differs from it only in that it cannot follow $x_t$.
As a control, we shuffle predictions across transitions while keeping the actual and alternative observations fixed, so that each prediction comes from an unrelated current observation.

In every tested setting, the systems identify the actual next observation well above chance, whereas shuffled predictions remain near chance (Table~\ref{tab:next-observation-prediction}).

\begin{table}[ht]
\centering
\caption{\textbf{Predictors distinguish actual next observations from incompatible alternatives.}
Values are percentage correct, reported as mean $\pm$ standard error across three training runs.
Chance is $50\%$ for choosing between the actual next observation and the alternative.
Each system is evaluated in its own learned latent space.
The side-information condition supplies the current $C$ image while keeping it visible in the main input.}
\label{tab:next-observation-prediction}
\begin{center}
\begin{tabular}{@{}lcccc@{}}
\hline
& \multicolumn{2}{c}{No side information} & \multicolumn{2}{c}{Side information} \\
\cline{2-3}\cline{4-5}
Setting & Prediction & Shuffled & Prediction & Shuffled \\
\hline
Traffic light ($q=1$) & $77.13\pm0.51$ & $49.97\pm0.43$ & --- & --- \\
\hline
Two pairs, $p_D=0.60$ & $88.88\pm0.35$ & $50.10\pm0.38$ & $91.27\pm0.19$ & $49.95\pm0.28$ \\
Two pairs, $p_D=0.75$ & $90.65\pm0.14$ & $50.27\pm0.23$ & $93.64\pm0.04$ & $50.13\pm0.11$ \\
Two pairs, $p_D=0.85$ & $92.22\pm0.08$ & $50.56\pm0.16$ & $95.39\pm0.07$ & $49.70\pm0.06$ \\
Two pairs, $p_D=0.95$ & $94.53\pm0.36$ & $49.83\pm0.33$ & $97.15\pm0.04$ & $49.59\pm0.07$ \\
Two pairs, $p_D=1.00$ & $96.99\pm0.49$ & $50.29\pm0.40$ & $97.96\pm0.13$ & $49.83\pm0.23$ \\
\hline
\end{tabular}
\end{center}
\end{table}

\section{Regularization Sensitivity}
\label{app:regularization-sensitivity}

\begin{figure}[H]
\centering
\includegraphics[width=0.96\linewidth]{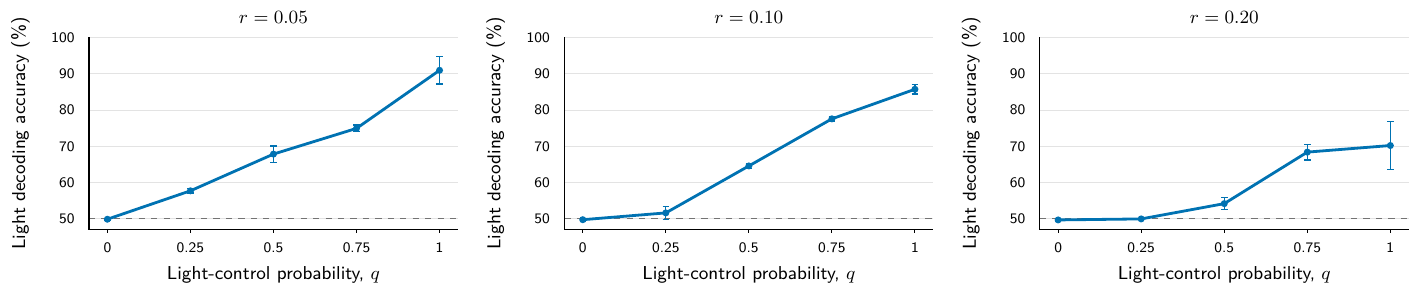}
\caption{\textbf{Cross-subsidization.}
The current light becomes more strongly represented as $q$ increases, although stronger regularization weakens this effect.
Dashed lines mark $50\%$ chance light-decoding accuracy.}
\label{fig:sensitivity-traffic}
\end{figure}

\begin{figure}[H]
\centering
\includegraphics[width=0.96\linewidth]{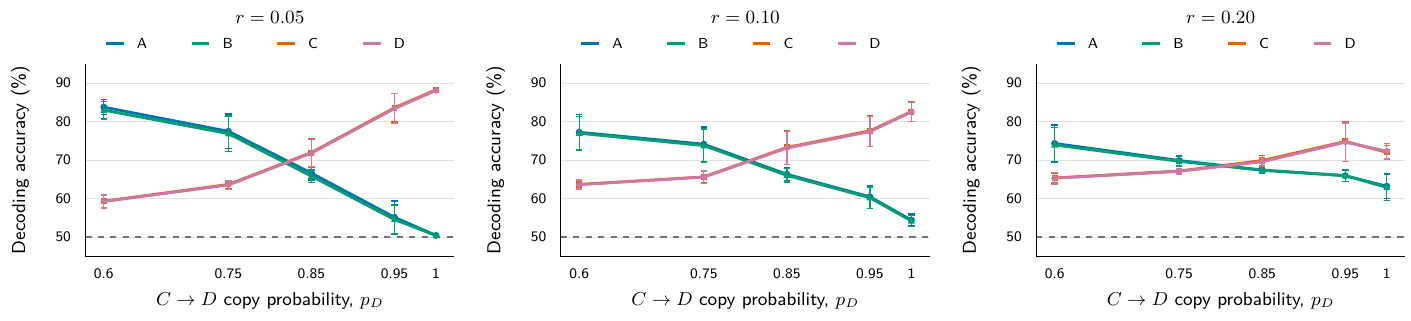}
\caption{\textbf{Changing predictability.}
Increasing $p_D$ from $0.60$ to $1$ changes the better-represented pair from $A/B$ to $C/D$ in all three runs at every regularization weight.
Stronger regularization reduces the separation between the pairs.
The change need not be monotonic: at $r=0.20$, mean decoding accuracy for $C/D$ decreases from $p_D=0.95$ to $p_D=1$.
Dashed lines mark $50\%$ chance accuracy.}
\label{fig:sensitivity-predictability}
\end{figure}

\begin{figure}[H]
\centering
\includegraphics[width=0.96\linewidth]{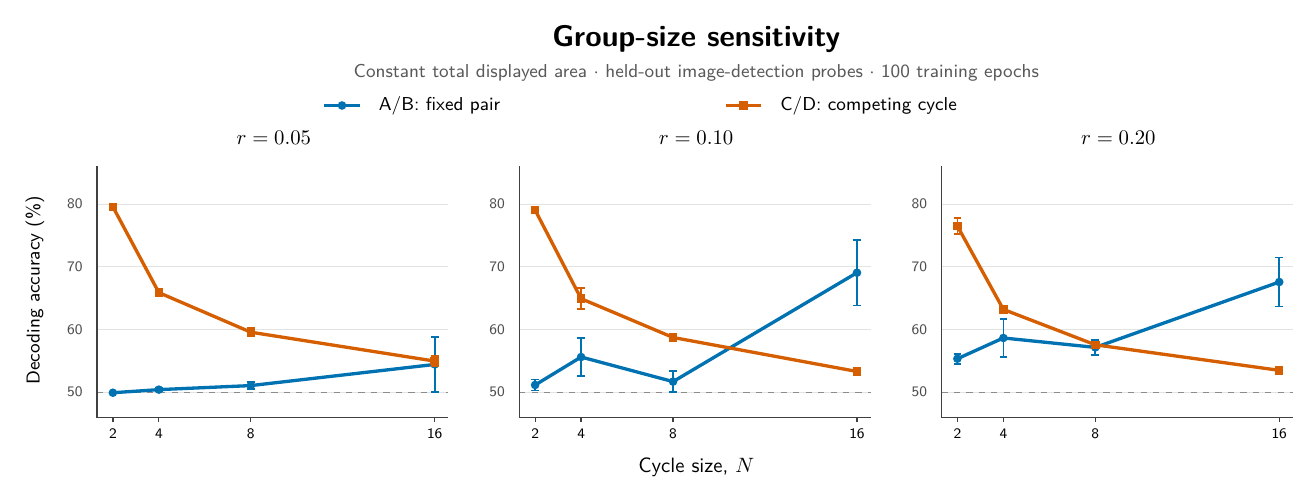}
\caption{\textbf{Changing group size.}
Enlarging the $C/D$ group reduces its decoding accuracy in every run at every regularization weight.
At $N=16$, $A/B$ becomes better represented in all three runs at $r=0.10,0.20$, but only one of three at $r=0.05$.
We average decoding accuracy across positions within each group; dashed lines mark $50\%$ chance accuracy.}
\label{fig:sensitivity-size}
\end{figure}

\begin{figure}[p]
\centering
\includegraphics[width=0.90\linewidth]{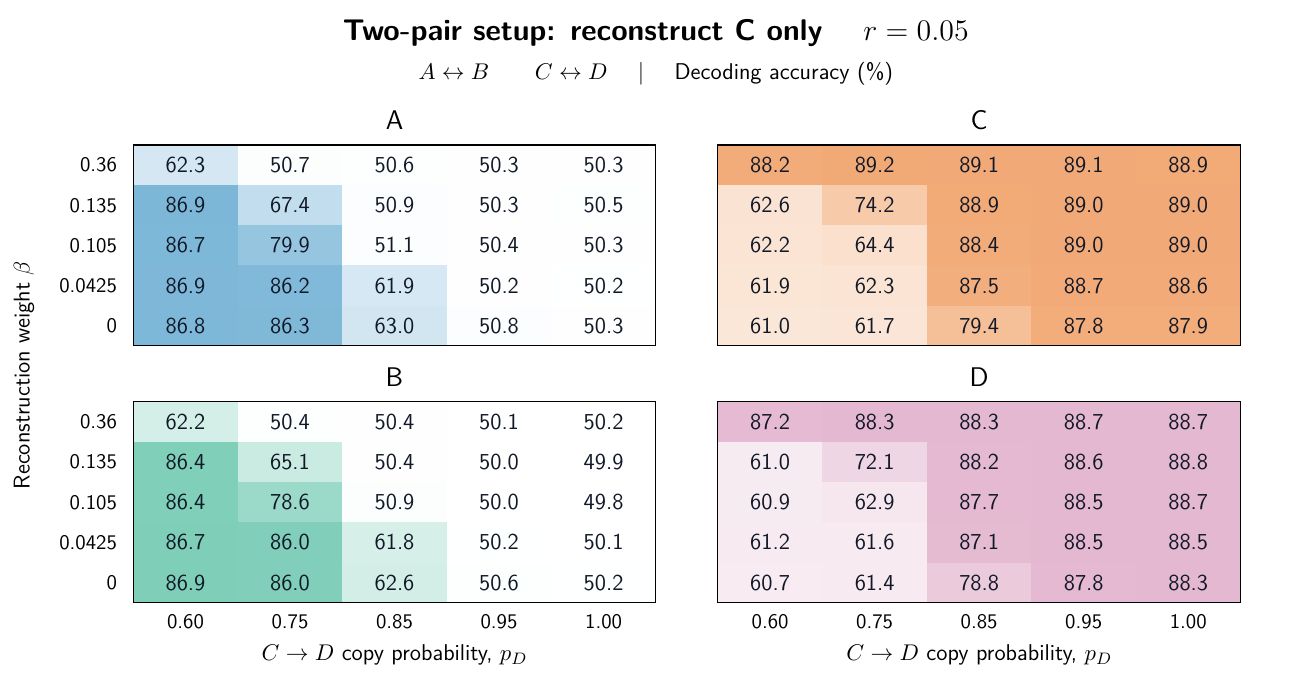}
\par\smallskip
\includegraphics[width=0.90\linewidth]{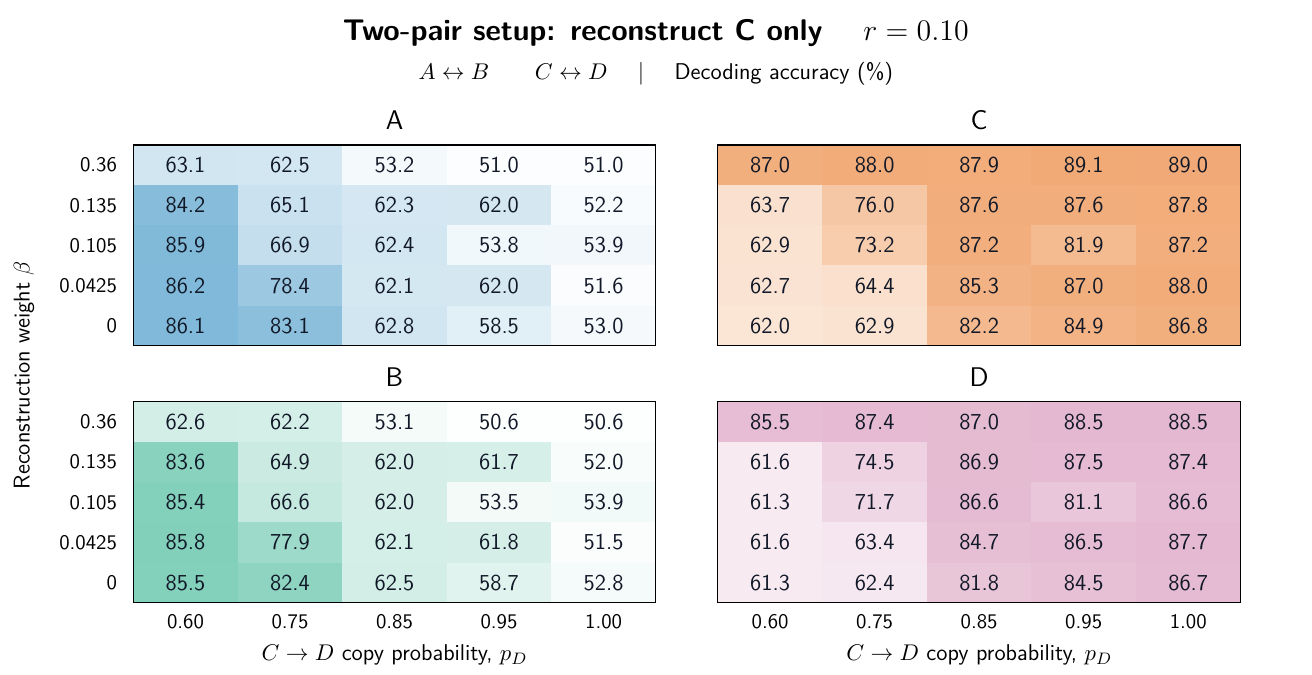}
\par\smallskip
\includegraphics[width=0.90\linewidth]{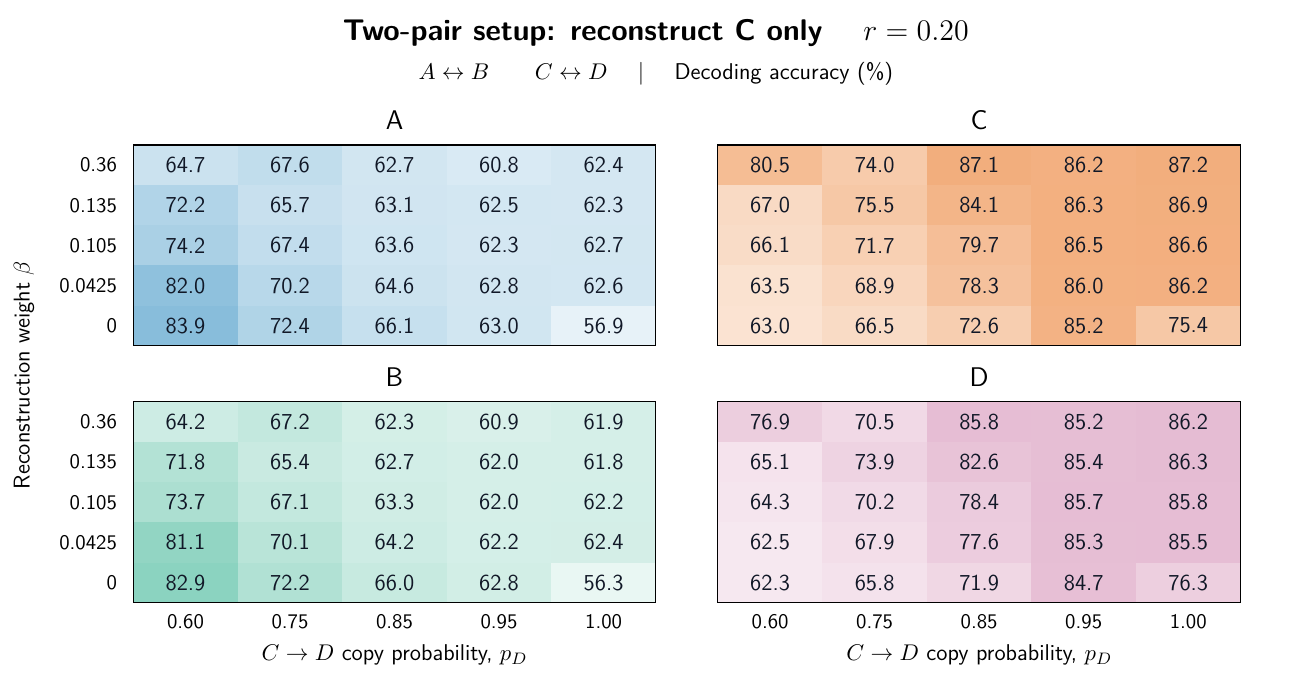}
\caption{\textbf{Auxiliary reconstruction.}
Reconstructing $C$ can also strengthen its predictive partner $D$, which receives no reconstruction supervision.
At $\beta=0.36$, $C/D$ is better represented than $A/B$ at every $p_D$ and regularization weight.
Increasing $\beta$ does not always improve decoding: at $r=0.20,p_D=0.75$, both $C$ and $D$ decline from $\beta=0.135$ to $\beta=0.36$.
All four images remain visible to the encoder.
Cells show decoding accuracy from one training run, using a shared $50$--$90\%$ color scale; numerical values below $50\%$ remain visible.}
\label{fig:sensitivity-rec-low}
\end{figure}

\begin{figure}[p]
\centering
\includegraphics[width=0.81\linewidth]{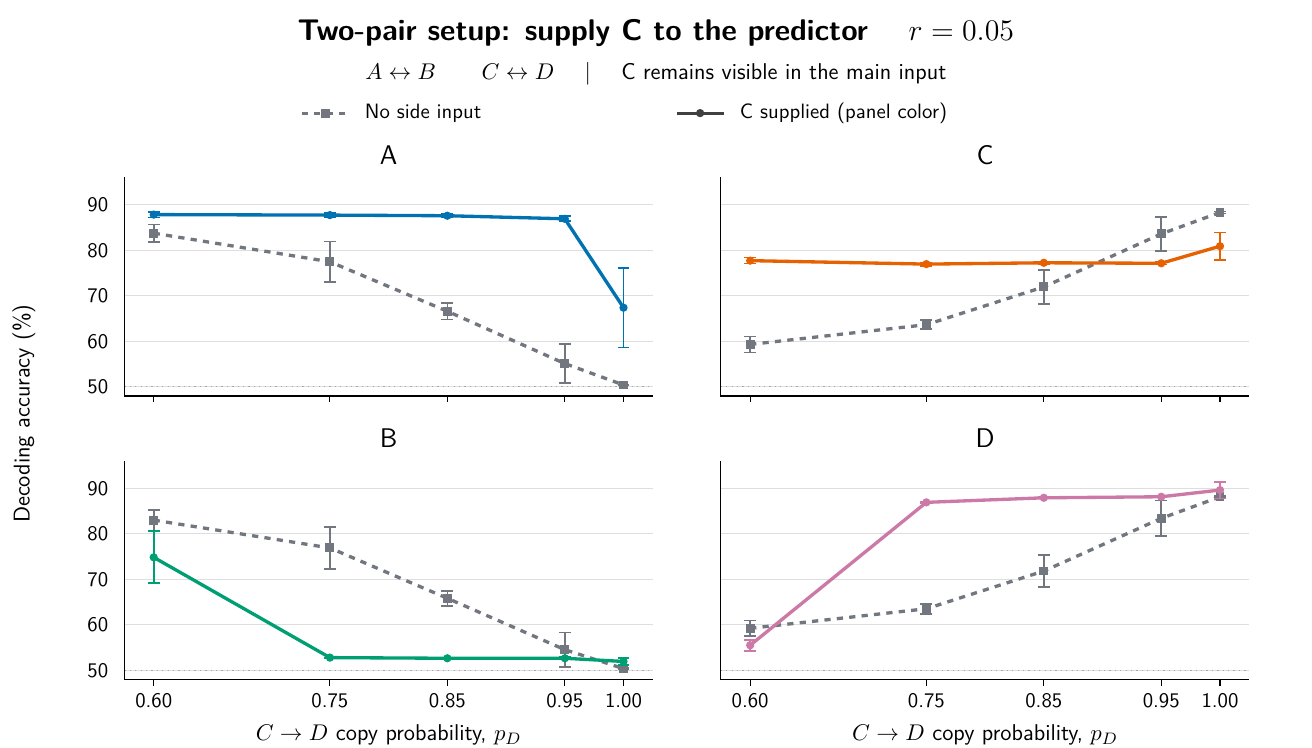}
\par\smallskip
\includegraphics[width=0.81\linewidth]{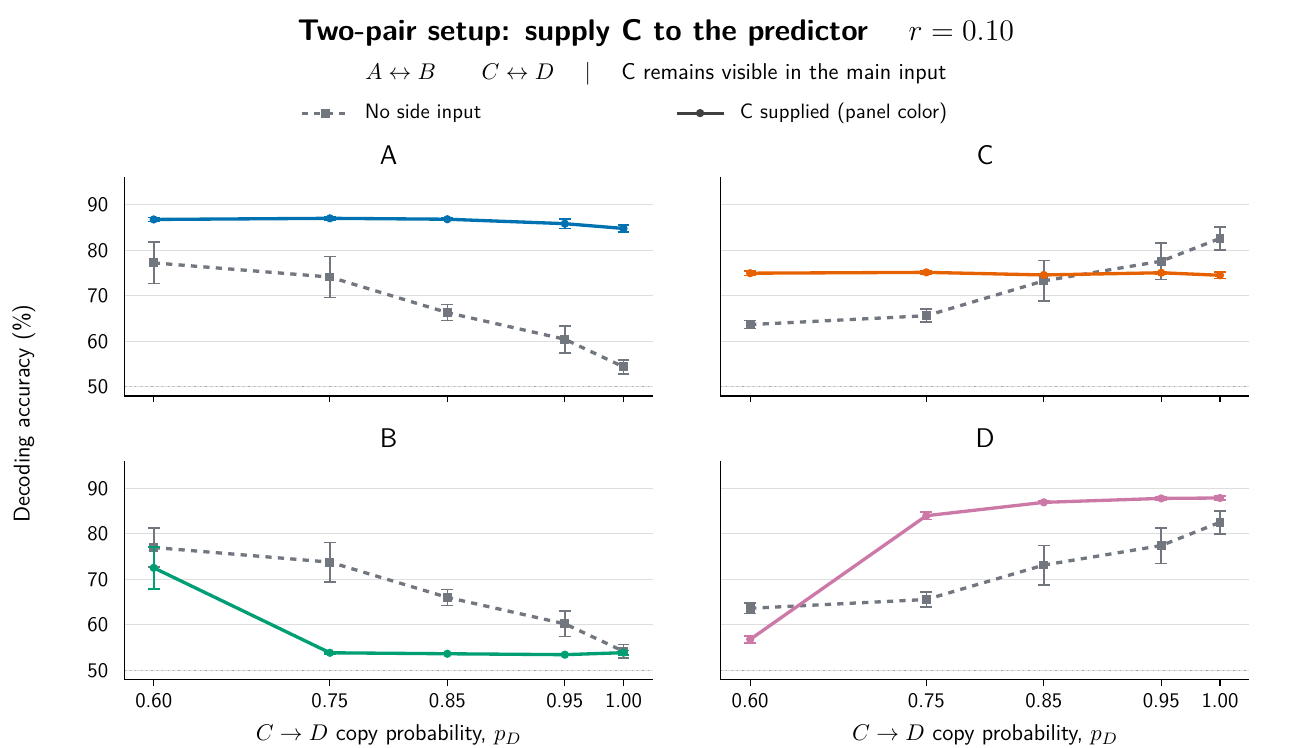}
\par\smallskip
\includegraphics[width=0.81\linewidth]{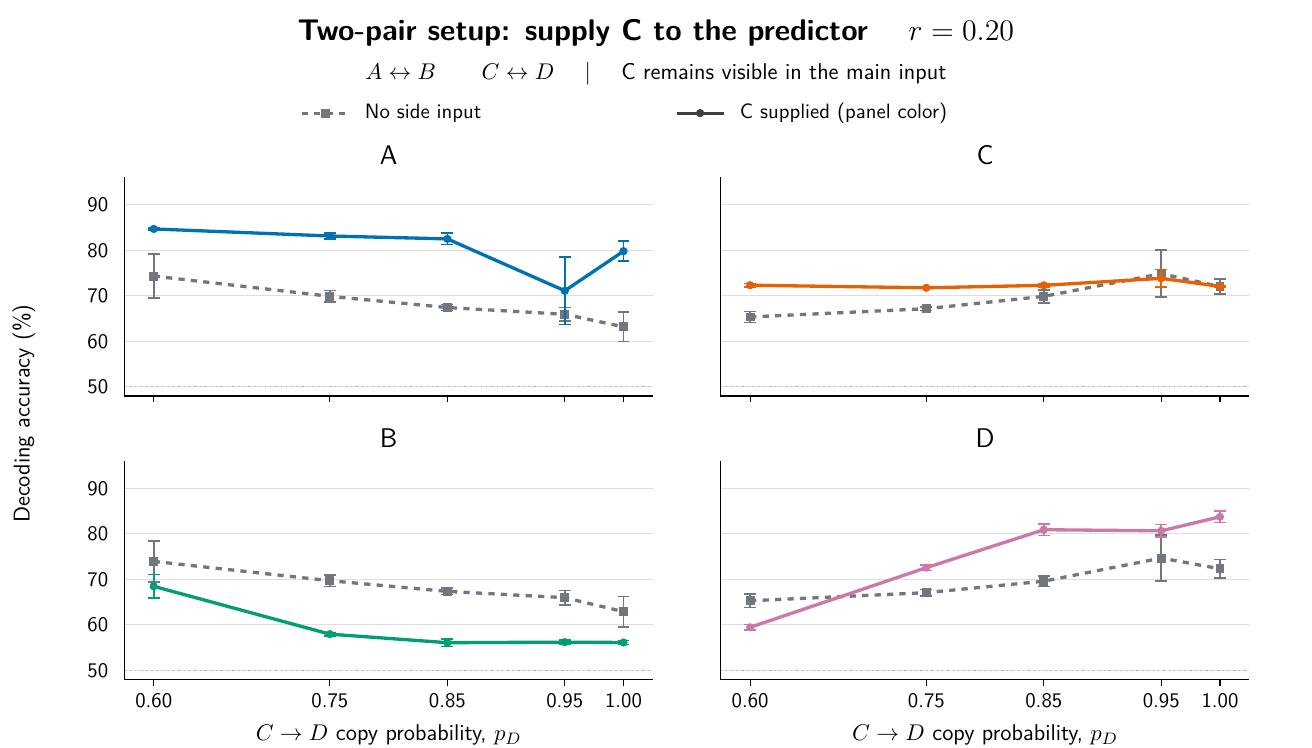}
\caption{\textbf{Predictor side information.}
Supplying $C$ strengthens representation of $A$ and $D$ while weakening $B$ at $p_D=0.75,0.85$ in all three runs at every regularization weight.
The effect varies across the sweep and does not uniformly replace the $A/B$ pair with $C/D$.
Colored curves supply $C$ to the predictor; gray curves omit side input.
All probes read the main latent without access to the side input; supplying $C$ also adds encoder capacity.
Panels show decoding accuracy with standard errors across three runs; dotted lines mark $50\%$ chance accuracy.}
\label{fig:sensitivity-side-low}
\end{figure}

We test how the findings change with the regularization weight in all five experiments, using
\begin{equation}
\mathcal{L}_r=(1-r)\mathcal{L}_{\mathrm{pred}}+r\mathcal{L}_{\mathrm{SIGReg}},
\qquad r\in\{0.05,0.10,0.20\}.
\label{eq:sensitivity-objective}
\end{equation}
The main setting uses $r=0.10$; the corresponding weights relative to a unit prediction coefficient are $\lambda=r/(1-r)$, about $0.05$, $0.11$, and $0.25$.
For reconstruction, we add $\beta\mathcal{L}_{\mathrm{rec},C}$ and hold $\beta$ fixed across $r$.
All other training and probing settings follow Appendix~\ref{app:experimental-details}.
Figures~\ref{fig:sensitivity-traffic}--\ref{fig:sensitivity-side-low} show every tested condition.
We report mean decoding accuracy with standard errors across three runs, except for reconstruction, which uses one run.

Across the tested $\lambda$ values, we recover the qualitative findings reported in the main text for cross-subsidization, predictability-driven set switching, and steering through reconstruction or side information.
The group-size result holds at the larger $\lambda$; at the smaller $\lambda$, the enlarged group still becomes less represented, but the switch in favor of the competing group occurs in only one of three runs.

\section{Worst- and Best-Position Readouts for Group Size}
\label{app:capacity-position-readouts}

We check whether the reversal in Figure~\ref{fig:set-size-switching} also appears at the worst and best positions, using the same frozen encoders and position-specific image-detection probes described in Appendix~\ref{app:set-setup-details}.
For each optimization seed, we first average accuracy across target identities at each position, then take the minimum or maximum across positions within each group.
Both measures show the same endpoint reversal in all three seeds: the competing cycle is more strongly represented at $N=2$, whereas the smaller $A/B$ group is more strongly represented at $N=16$.
The average probability of copying an image from the preceding position in each cycle remains unchanged across these sizes.

\begin{figure}[H]
\centering
\includegraphics[width=0.88\linewidth]{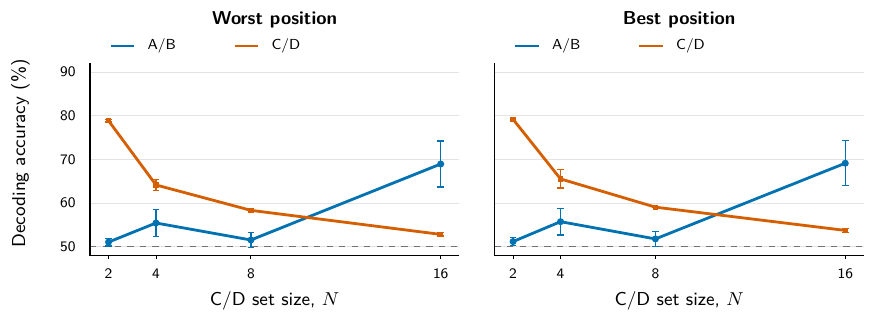}
\caption{\textbf{Worst- and best-position readouts show the same reversal.}
Both measures favor the competing cycle at the smallest size and $A/B$ at the largest, showing that the reversal also holds for the least and most decodable positions within each group.
Curves show means across three training runs; error bars show standard errors.
Probes are trained on collages from images held out from encoder training and evaluated on separate collages.
The dashed line marks the chance image-detection accuracy of $50\%$.}
\label{fig:capacity-extreme-positions}
\end{figure}

$C/D$ accuracy decreases across all four sizes in every seed, but $A/B$ accuracy is not monotonic, and the intermediate-size preferences vary across seeds.
At $N=16$, $A/B$ decoding also varies more across optimization seeds.
We hold the training and evaluation budget per position fixed across group sizes.
Appendix~\ref{app:set-setup-details} describes the constant-area rendering procedure.

\end{document}